\documentclass{article}

\PassOptionsToPackage{numbers, compress}{natbib}

\usepackage{lipsum}
\usepackage{amsmath}
\usepackage{dsfont} 
\usepackage{graphicx}
\usepackage{subfigure}
\usepackage{bm}
\usepackage{amsmath}
\usepackage{amssymb}
\usepackage{mathtools}
\usepackage{amsthm}
\usepackage{enumitem}
\theoremstyle{plain}

\theoremstyle{definition}

\theoremstyle{remark}

\newcommand{\cR}{\mathbb{R}}

\newcommand{\ours}{CogRobot}

\usepackage{algorithm}
\usepackage{algorithmic}

\usepackage[preprint]{neurips_2025}

\usepackage{footnote}
\usepackage[utf8]{inputenc} 
\usepackage[T1]{fontenc}  
\usepackage{hyperref}       
\usepackage{url}     
\usepackage{booktabs}      
\usepackage{amsfonts}     
\usepackage{nicefrac}      
\usepackage{microtype}      
\usepackage{xcolor}     
\usepackage{wrapfig}

\usepackage{colortbl}
\usepackage{xcolor}
\usepackage{pifont}

\usepackage{tabularx}
\usepackage{booktabs}
\usepackage{multirow}
\usepackage{arydshln} 

\title{Towards a Generalizable Bimanual Foundation Policy via Flow-based Video Prediction}

\author{%
\textbf{Chenyou Fan$^{\,1,2}$\thanks{Equal Contribution}
$\quad$ Fangzheng Yan$^{\,1,3*}$ $\quad$ Chenjia Bai$^{1}$\thanks{Project lead} $\quad$ Jiepeng Wang$^{1}$} \\ $\quad$ \textbf{Chi Zhang$^{1}$ $\quad$ Zhen Wang$^{2}$ $\quad$ Xuelong Li$^{1}$}
  \\\\
$^1$Institute of Artificial intelligence (TeleAI), China Telecom \quad \\
$^2$Northwestern Polytechnical University \quad  
$^3$Hong Kong University of Science and Technology
}

\begin{document}

\maketitle

\begin{abstract}

Learning a generalizable bimanual manipulation policy is extremely challenging for embodied agents due to the large action space and the need for coordinated arm movements. Existing approaches rely on Vision-Language-Action (VLA) models to acquire bimanual policies. However, transferring knowledge from single-arm datasets or pre-trained VLA models often fails to generalize effectively, primarily due to the scarcity of bimanual data and the fundamental differences between single-arm and bimanual manipulation. In this paper, we propose a novel bimanual foundation policy by fine-tuning the leading text-to-video models to predict robot trajectories and training a lightweight diffusion policy for action generation. Given the lack of embodied knowledge in text-to-video models, we introduce a two-stage paradigm that fine-tunes independent text-to-flow and flow-to-video models derived from a pre-trained text-to-video model. Specifically, optical flow serves as an intermediate variable, providing a concise representation of subtle movements between images. The text-to-flow model predicts optical flow to concretize the intent of language instructions, and the flow-to-video model leverages this flow for fine-grained video prediction. Our method mitigates the ambiguity of language in single-stage text-to-video prediction and significantly reduces the robot-data requirement by avoiding direct use of low-level actions. In experiments, we collect high-quality manipulation data for real dual-arm robot, and the results of simulation and real-world experiments demonstrate the effectiveness of our method. 
\end{abstract}

\section{Introduction}
\label{sec:intro}

Bimanual manipulation represents a pivotal for embodied agents, enabling them to perform complex tasks that require precise coordination between two arms. Unlike single-arm manipulation, bimanual tasks necessitate human-like coordination, which introduces significant challenges due to the substantially larger action space and the requirement for coordinated movements of two arms. Prior approaches have leveraged adopt simulation \cite{grotz2024peract,mu2024robotwin} or small-scale real-world data \cite{Aloha,liu2024voxactb,grannen2023stabilize,franzese2023interactive}, human-objective primitives \cite{gao2024bi,liu2024taco}, and reinforcement learning (RL) \cite{chen2022towards,lin2023bi} with policy transfer to learn a bimanual policy. However, these methods often exhibit limited generalization capabilities due to the scarcity of high-quality bimanual data, and the sim-to-real gap remains a significant challenge for RL-based approaches. These challenges necessitate the development of a bimanual foundation policy that can generalize across tasks, while maintaining adaptability and resilience in the real world. 

Recent approaches construct large-scale Vision-Language-Action (VLA) models to acquire generalizable bimanual policies. Several methods direct mixing cross-embodied data to train a single network, while the data mixing strategy is proven to be important for downstream adaptation, particularly when domain-specific robotic data are limited \cite{openvla,pi0}. Other research attempts to build a unified action space for all robots, enabling joint pre-training on single- and dual-arm datasets alongside web-scale human data. Specifically, RDT \cite{rdt} introduces a physically interpretable action space, where each dimension maintains explicit correspondence to physical primitives, thereby facilitating transferability across robotic platforms. GROOT and GO-1 \cite{GR00T,GO1} employ latent action embeddings to project different action spaces into a shared codebook, enabling large-scale pre-training on heterogeneous data. Despite these innovations, training a robust VLA model for bimanual manipulation remains challenging. First, these frameworks necessitate training from scratch based on aggregated datasets with unified action space, which introduces substantial computational overhead. Second, even within a unified action space, bimanual manipulation exhibits pronounced multi-modality since it inherently contains many possible action modes \cite{rdt}, requiring comprehensive data coverage. 

The challenges in building a generalist bimanual VLA raise a key question: Is it feasible to construct a bimanual policy using a foundation model without directly handling heterogeneous actions? Observations suggest that despite the complexity of bimanual manipulation actions, the resulting state trajectories can be uniformly represented through videos. Thus, establishing a unified modeling framework for both natural and robotic videos conditioned on language instructions could enable a bimanual foundation model to predict robot trajectories from instructions. Fortunately, the existing text-to-video (T2V) models \cite{sora,cogvideox} have demonstrated remarkable generation and instruction-following capabilities.  These models capture rich motion semantics, including human interactions with the physical world, which are crucial for embodied agents to model robot and object dynamics. Additionally, T2V models inherently maintain temporal dependencies, making them suitable for modeling trajectories in Markov Decision Processes (MDPs). 

In this paper, we introduce \textbf{\ours}, which leverages the state-of-the-art T2V model, CogVideoX~\cite{cogvideox}, to derive a bimanual foundation policy. By fine-tuning CogVideoX on a bimanual dataset, we can adapt the T2V model to effectively predict future robot trajectories based on initial observation and language instruction, eliminating the need for training from scratch or modeling complex actions as in previous VLA models. A lightweight diffusion policy then generates robot actions from the predicted videos. However, fine-tuning CogVideoX directly on limited bimanual data often yields suboptimal predictions. We attribute this to the model's lack of embodied domain knowledge, hindering precise prediction of robotic and object movements. To address this, \ours~introduces optical flow as an intermediate variable, decomposing the process into two modules: \emph{text-to-flow} and \emph{flow-to-video}, both built upon CogVideoX. (i)~Compared to direct prediction videos, the \emph{text-to-flow} module predicts optical flow rather than raw videos, focusing on modeling kinematic behavior and dynamic interactions of robots and objects without detailing video specifics. (ii)~Then, the \emph{flow-to-video} reconstructs detailed videos from the flow, grounding language instructions in concrete high-level semantics via the flow without considering low-level actions. The \ours~framework alleviates ambiguity in language instructions and significantly reduces data requirements for fine-tuning. \ours~thus efficiently bridges high-level instructions with low-level actions through intermediate physical representations. 

Our contributions are threefold: (i) We propose a framework for training a bimanual foundation policy by leveraging T2V models. (ii) We introduce a two-stage paradigm that employs optical flow as concise video representations, thereby reducing the data requirements for bimanual manipulation. (iii) We build a dual-arm platform to collect high-quality bimanual manipulation data and evaluate our method in both simulation and real robot. Results demonstrate that \ours~accurately predicts robot trajectories after fine-tuning the T2V model on limited data. The resulting policy outperforms baselines in various challenging tasks and exhibits strong generalization to unseen scenarios. 

\section{Preliminaries}

We adopted two 7-DoF Realman robotic arm and an external camera to build a dual-arm system, as shown in Fig.~\ref{fig:framework}(a). The bimanual manipulation task $\mathcal{T}$ can be formulated as a goal-conditioned Partially Observable Markov Decision Process (POMDP) parameterized by $(\mathcal{S}, \mathcal{O}, \mathcal{A}, P, \mathcal{L})$ where $\mathcal{S}$ and $\mathcal{O}$ are state and observation space, respectively. In this work, we consider the observation $o_t\in \cR^{H\times W\times 3}$ as the RGB image captured by an external camera, where $H$ and $W$ are the height and width of the image. The action $a_t$ is the combined joint positions of the two arms, which can be directly applied to control the robot. $P:\mathcal{S}\times\mathcal{A}\rightarrow\mathcal{S}$ is the transition function, $R:\mathcal{S}\times\mathcal{A}\rightarrow \{0,1\}$ is a binary reward function that measures whether the language goal described by $l$ is achieved. We collect expert data through teleoperation of the VR device, which captures human hand and wrist poses and translate them into robot joint angle command through re-targetting and inverse kinematics, basically following Open-Television \cite{opentelevision}. The collected dataset contains episodic data that include observation sequences (i.e., video) $v=[o_1,\ldots,o_T]\in \cR^{T\times H\times W\times 3}$, action sequences $[a_1,\ldots,a_{T-1}]$, and language description $l$.

\section{Methods}
\label{sec:methods}

In CogRobot, we decompose the training of the instruction-conditioned bimanual policy $\pi(a_{t:t+N-1} | o_t, l)$ into two steps: (\romannumeral1) predicting future observation trajectories $o_{t+1:t+N} = { o_{t+1}, \ldots, o_{t+N} }$ that achieve the specified goal $l$, given the current observation $o_t$; and (\romannumeral2) deriving executable low-level actions $a_{t:t+N-1}$ from the predicted observation sequences $o_{t+1:t+N}$. In this paper, the prediction of $o_{t+1:t+N}$ is formulated as a video generation problem.

\begin{figure*}[t]
    \centering
    \includegraphics[width=.95\textwidth]{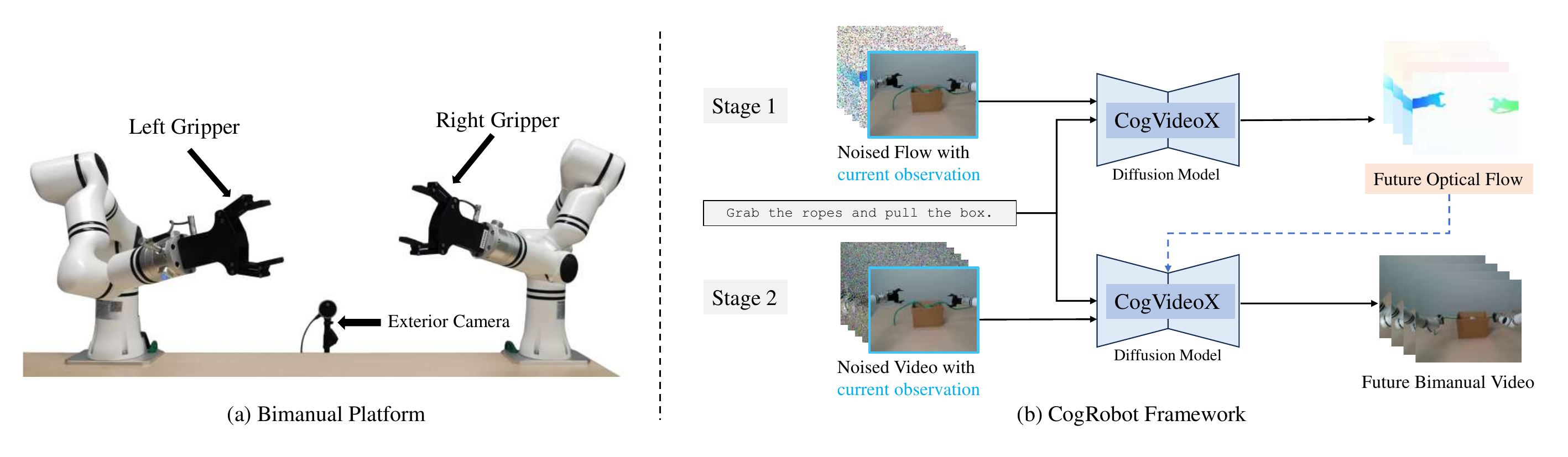}
    \caption{Overview of {\ours}. (a) We build a dual-arm robotic system based on Realman arms and use a single-view RGB observation captured by an Intel RealSense camera. (b) We propose a two-stage pipeline for bimanual trajectory prediction. The first stage focuses on modeling potential arm motion, which is then used as additional guidance for the second-stage video prediction.}
    \label{fig:framework}
    \vspace{-1.5em}
\end{figure*}

Recent T2V models have demonstrated strong capabilities in generating highly realistic videos by training on large-scale open-domain text-video datasets.
However, due to the absence of bimanual manipulation data in these datasets, such models struggle to generalize to dual-arm robotic systems to forecast their future behaviors. 
Fine-tuning on downstream tasks is necessary but presents two challenges:
(1) Dual-arm systems require coordinated actions between both arms, which introduces more complex dynamics that are difficult for existing T2V models to represent. 
(2) Publicly available dual-arm datasets are scarce due to the high cost of data collection. 
As a result, directly fine-tuning existing video diffusion models often leads to inaccurate predictions of arm movements and fails to capture the dynamic interactions of robots and objects, resulting in unreliable trajectories for bimanual execution.
To overcome these challenges, we propose a two-stage fine-tuning framework that explicitly integrates motion details into the video generation process, as illustrated in Fig.~\ref{fig:framework}(b). We leverage optical flow to encode pixel-level motion and predict future optical flow sequences that reflect fine-grained motion patterns. These predicted flows are used to guide the video generation process, enabling the model to produce more precise videos.


\subsection{Text-to-Flow Generation}
\label{sec:t2fmodel}

\begin{figure*}[t]
    \centering
    \includegraphics[width=\textwidth]{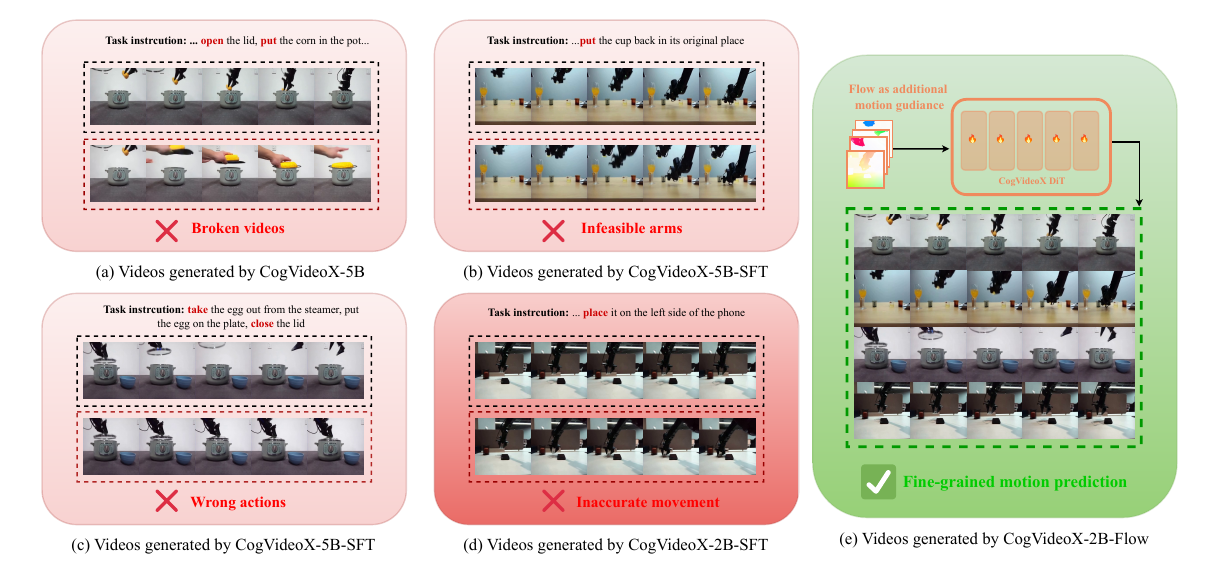}
    \vspace{-20pt}
    \caption{Visualization of different models. (a) Videos are generated directly using the CogVideoX-5B without any fine-tuning. (b) and (c) show videos from the CogVideoX-5B directly fine-tuned on bimanual datasets RDT and RoboMIND. (d) presents videos from the CogVideoX-2B directly fine-tuned on the same datasets. (e) shows videos from the CogVideoX-2B fine-tuned with optical flow. The ground truth videos are shown in the black dashed box while the generated videos are shown in the red dashed box in (a)-(d) and the green dashed box in (e).}
    \label{fig:method_example}
    \vspace{-1.5em}
\end{figure*}





We first demonstrate the limitations of current T2V models in predicting bimanual behaviors. We adopt CogVideoX as the base model and evaluate its zero-shot generalization capability using two dual-arm datasets: RDT \cite{rdt} and RoboMIND \cite{robomind}. We then fine-tune CogVideoX on a combined dataset comprising these two datasets to assess its ability to adapt to dual-arm systems.

As shown in Fig.~\ref{fig:method_example}(a), when given the initial observation and prompted with the task "open the lid and put the corns in the pot", the vanilla CogVideoX generates a spurious human hand and a corn to follow the instruction. It fails to incorporate the visual semantics of the dual-arm in the initial observation and instead focuses solely on producing frames that align with the text prompt. As a result, it is unable to plan valid trajectories.

We fine-tune two versions of CogVideoX with different model sizes (2B and 5B) on the same dataset, and refer to them as CogVideoX-2B-SFT and CogVideoX-5B-SFT. As shown in Fig.~\ref{fig:method_example}(b)-(d), directly fine-tuning on the bimanual dataset leads to inferior performance due to semantic ambiguity introduced by high-level task instructions. Specifically, it can be attributed to three challenges:
(\romannumeral1) \textbf{Physical hallucination}. As shown in Fig.~\ref{fig:method_example}(b), even after fine-tuning, the hallucination observed in Fig.~\ref{fig:method_example}(a) persists—the generated video replaces the human hand with a fictitious robotic arm, but the motion still violates physical realism. This suggests that T2V models primarily optimize for generating videos that are semantically aligned with the textual input, without ensuring that each frame follows physically plausible transitions.
(\romannumeral2) \textbf{Task confusion}. When prompted with a long-horizon instruction involving multiple stages, the model may fail to determine the correct subtask to execute at a given time. As shown in Fig.~\ref{fig:method_example}(c), when different subtasks share similar initial observations, the model cannot infer whether the next step should be to 'take out from' or 'close', and may incorrectly trigger an unintended subtask.
(\romannumeral3) \textbf{Vague instructions}. The language instruction is always high-level (e.g., "place it") and lacks sufficient guidance on how the action should be executed. As a result, as shown in Fig.~\ref{fig:method_example}(d), while the generated videos may appear to fulfill the instruction, the lack of precision leads to aggressive movement plans that cannot be executed in future $N$ steps. The model may produce implausible joint positions that are not grounded in feasible robot dynamics.
These challenges highlight a persistent gap between textual descriptions and real-world motion, which hinders the model’s ability to acquire the necessary knowledge to predict how the arms should move and interact with the target objects over time.

To this end, we propose to leverage optical flow as guidance to explicitly provide the model with motion cues. Specifically, given a pair of RGB observations, $o_1$ and $o_2$, optical flow computes a per-pixel displacement field $\mathbf{f}_{1 \to 2} : \mathbb{R}^2 \to \mathbb{R}^2$ that maps each pixel location $\mathbf{p} \in \mathbb{R}^2$ in observation $o_1$ to its corresponding location $\mathbf{p}' = \mathbf{p} + \mathbf{f}_{1 \to 2}(\mathbf{p})$ in the future observation $o_2$. In bimanual manipulation, the differences between two image observations are primarily caused by the motion of the robotic arms and the interaction with target objects. Therefore, by modeling pixel-wise variations, optical flow can capture fine-grained motion details of arm joints and end-effectors, as well as the contacts between the robot and objects. This capability allows for precise monitoring of both the kinematic behavior of the robotic arm and the dynamic interactions at the contact interface. To enhance temporal consistency across flow predictions at different future time steps, we formulate the flow prediction process as a sequence modeling problem. Specifically, the objective of the first stage is to generate a sequence of optical flow $\mathbf{F}_{0:N} = (\mathbf{f}_{0 \to 1}, \dots, \mathbf{f}_{0 \to N}) \in \mathbb{R}^{N \times H \times W \times 2}$, where each $\mathbf{f}_{0 \to t}$ represents the motion field from the initial observation $o_0$ to the future observation $o_t$.

However, directly predicting raw optical flow introduces new challenges. Due to the modality gap between 2-channel optical flow and 3-channel RGB images, previous work \cite{motioni2v} requires training an additional flow Variational Autoencoder (VAE) from scratch to transfer the model from the RGB domain to the motion vector domain. This approach relies on web-scale training data to effectively adapt the model’s distribution, which is impractical for bimanual manipulation tasks due to the limited availability of domain-specific data. To better leverage the pretrained model’s prior and reduce dependence on large-scale data, we instead convert optical flow into a flow video format. Specifically, for each optical flow $\mathbf{f}=(u,v)$, we calculate its magnitude and angle as follows:
\begin{equation}
\| \mathbf{f} \| = \sqrt{u^2 + v^2}, \quad \theta = \operatorname{atan2}(-v, -u).
\end{equation}
Different flow directions are mapped to distinct colors chosen from the color wheel used in \cite{raft}, while the magnitude of the flow determines the saturation of the color. By iteratively converting each optical flow in the sequence $\mathbf{F}_{0:N}$, we can generate a flow video $v_{\mathbf{F}} = (v_{\mathbf{f}_{0 \to 1}}, \dots, v_{\mathbf{f}_{0 \to N}}) \in \mathbb{R}^{N \times H \times W \times 3} $ that preserves the ability to capture arm motion while eliminating the format gap. Since this conversion process is rule-based and involves straightforward calculations, it is highly efficient and introduces minimal computational overhead compared to the cost of model training.

With this transformation, the flow generation task is reformulated as learning a distribution over flow videos: $v_{\mathbf{F}} \sim p_{\theta_{f}}(v_{\mathbf{F}} \mid o_0, l). $
To enable effective training with a limited dual-arm dataset, we adopt the leading T2V model CogVideoX as our base model, leveraging its generative priors learned from large-scale web videos. During training, we initialize the model with pretrained weights and freeze the VAE to encode the flow video $v_{\mathbf{F}}$ into a latent $z_f^0$. This latent is then perturbed with progressively added noise $\{\epsilon_f^k\}_{k=0}^{K}$. At each denoising step $k$, the model receives the noisy latent $z_f^k$ and predicts the corresponding noise $\epsilon_f^k$. The loss function for flow generation is defined as:
\begin{equation}
    \label{eq:flow_loss}
    \mathcal{L}_{\text{flow}}(\theta_f) = \mathbb{E}_{k, \epsilon,z_f^0, o, l} \left[ \left\| \epsilon_f - \epsilon_{\theta_f}(z_f^k, k, o, l) \right\|^2 \right].
\end{equation}

\subsection{Flow-to-Video Prediction}




\begin{figure*}[t]
    \centering
    \includegraphics[width=\textwidth]{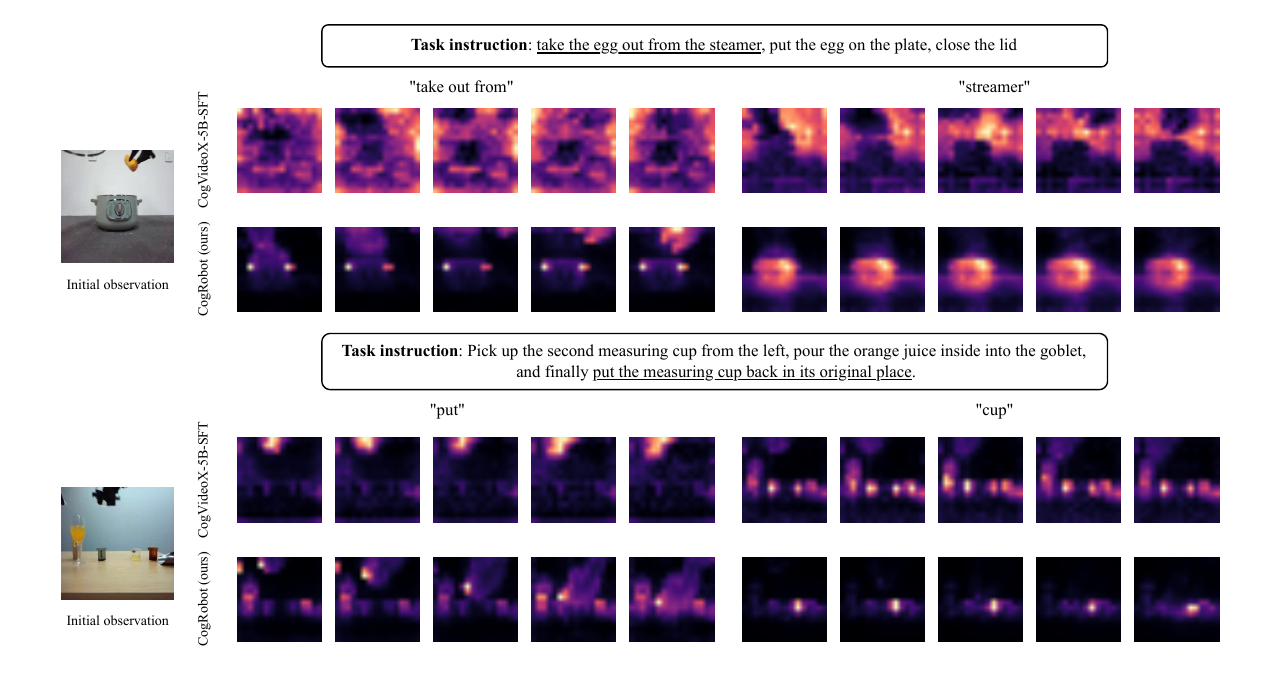}
    \vspace{-30pt}
    \caption{Attention maps across different models. The results are obtained using same initial observation and task instruction. The subtask corresponding to the current video clip is underlined.}
    \label{fig:attn_map}
    \vspace{-1em}
\end{figure*}

Based on the text-to-flow model, we build a flow-to-video model that leverages the motion and interaction cues embedded in optical flow to guide video synthesis. These cues capture dynamic patterns that are difficult to infer from text alone. In contrast to existing world models that rely on low-level action inputs, optical flow provides higher-level semantic information, which is helpful for identifying arms and objects. 
To demonstrate this, we visualize the attention maps of two models: CogVideoX-5B-SFT and our model, which incorporates optical flow as guidance. We select specific words from the instruction and extract the corresponding cross-attention maps to assess whether the models can correctly interpret these words and translate them into relevant low-level visual details. As shown in Fig. \ref{fig:attn_map}, for words related to arm motions, the language-only model fails to identify meaningful regions, leading to weak motion grounding. In contrast, our flow-guided model can respond to the expected motion and relevant target objects, achieving better alignment between the instruction and the visual output.
For words related to target objects, our model accurately segments the relevant regions and effectively filters out irrelevant background content, whereas the language-only model introduces more noise and highlights incorrect areas.

The model aims to learn a distribution over robotic trajectory videos with additional flow guidance:
$v \sim p_{\theta_{v}}(\cdot \mid o_0, l, v_{\mathbf{F}}). $
To effectively incorporate the flow video generated in the first stage into the video generation process, we propose concatenating the optical flow and the RGB video along the channel dimension. Specifically, during training, the flow video $v_{\mathbf{F}}$ and the dual-arm trajectory video $v$ are separately encoded by the VAE to obtain the corresponding latent $z_f$ and $z_v^0$, respectively. Following the same procedure used in flow generation, the video latent $z_v^0$ is perturbed by noise $\{\epsilon^k_v\}_{k=0}^{K}$ to produce $z_v^k$. This noisy latent is then concatenated with the flow latent to form $z^k=[z_v^k, z_f]$, which is fed into the model. The training objective for video generation is defined as:
\begin{equation}
    \label{eq:video_loss}
    \mathcal{L}_{\text{video}}(\theta_v) = \mathbb{E}_{k, \epsilon_v, o, l, (z_v^0,z_f)} \left[ \left\| \epsilon_v - \epsilon_{\theta_v}(z^k, k, o, l) \right\|^2 \right].
\end{equation}

\subsection{Diffusion Policy from Videos}



To enable the use of the video prediction model for controlling the dual-arm system, we introduce an additional controller that extracts executable low-level actions from the predicted videos. Since the video prediction model is trained to generate future observations that fulfill the given task instruction, each frame in the predicted video can be treated as a target observation that the dual-arm system should reach. Accordingly, we train a goal-reaching policy $\pi(\bm{a}_{0:n}|o_0, o_n)$ where $o_0$ is the initial observation and $o_n$ is the desired observation that the policy aims to achieve. We adopt Diffusion Policy \cite{diffusionpolicy} as the backbone for this controller. During training, we randomly sample a goal step $n \sim \mathcal{U}(1, n_{\max})$ and progressively perturb the corresponding action sequence $\bm{a}_{0:n}$ into $\bm{a}_{0:n}^k$ using noise $\{\epsilon^k_{\pi}\}_{k=0}^{K}$. The training objective of the goal-reaching policy is defined as:
\begin{equation}
    \mathcal{L}_{\text{policy}}(\theta_{\pi}) = \mathbb{E}_{k, \epsilon_{\pi}, (\bm{s}_0, \bm{a}^0_{0:n}), n \sim \mathcal{U}(1, n_{\max})} \left[ \left\| \epsilon_{\pi} - \epsilon_{\theta_{\pi}}(\bm{a}_{0:n}^k, \bm{s}_0, \bm{s}_n, k) \right\|^2 \right].
\end{equation}

\section{Related Works}

\paragraph{Bimanual Manipulation} Bimanual manipulation presents distinct challenges compared to single-arm setups, constrained by data scarcity \citep{lioutikov2016learning,stepputtis2022system}, expanded action spaces \citep{Aloha}, diverse collaboration modalities \citep{xie2020deep,franzese2023interactive}, and limitations in simulation fidelity or cost-effective real-world interfaces. Recent efforts address these issues through dual-arm simulation benchmarks \cite{humanoidbench,bigym,mu2024robotwin}, low-cost teleoperation systems for data acquisition \cite{Aloha,mobilealoha}, or data augmentation techniques \cite{mimicgen,dexmimicgen}. However, these approaches are constrained by hardware-specific limitations or insufficiently address the sim-to-real gap in policy generalization \cite{chen2022towards,lin2023bi}. Alternative methods focus on learning human-object interaction primitives \cite{gao2024bi,liu2024taco}, extracting geometric constraints via keypoint analysis \cite{Bikvil}, parameterizing movement primitives \cite{interactive}, or modeling dual-arm dependencies through attention mechanisms \cite{interactACT}. Although effective in specialized contexts, these techniques often depend on domain-specific feature extraction tools and strong prior assumptions, limiting their applicability to capture the full spectrum of action modes inherent in bimanual manipulation.

\paragraph{Foundation Model for Manipulation} Training foundation models holds promise for developing generalizable embodied policies. VLA architectures harness common sense knowledge from large-scale visual language models (VLM) to interpret language instructions and image observations, subsequently predicting robot actions through prediction heads or specialized tokens \cite{openvla,octo} when trained on cross-embodied datasets. To address the heterogeneous action spaces of diverse robotic platforms, particularly in bimanual manipulation, several approaches construct unified action spaces \cite{rdt}, latent action embeddings \cite{GR00T,GO1}, or spatial action grids \cite{spatialvla}, allowing large-scale pre-training on heterogeneous data. Other works address the continuous, multimodal nature of physical actions through diffusion-based action prediction \cite{cogact,dexgraspvla,molevla,he2024learning}. However, these methods typically require training from scratch on aggregated datasets, which incurs significant computational overhead. Additionally, the scarcity of bimanual manipulation data requires complex data mixing or extensive data collection to ensure comprehensive coverage of various action modes \cite{pi0,rdt}.

\paragraph{Text-to-Video Generation} Recent advances in text-to-video (T2V) generation have demonstrated significant improvements \cite{vdm,cogvideo}. Sora \cite{sora} leverages a diffusion-transformer backbone (DiT) to achieve high-resolution and long-duration video synthesis. CogVideoX \cite{cogact} further enhances performance in generating temporally consistent long-term videos, offering promising potential for robot video generation with appropriate fine-tuning. Previous methods utilizing T2V models for robot video generation \cite{UniPi,learn-to-act,grounding-video,robodreamer} typically rely on small-scale pre-trained models, which limits their capabilities to relatively straightforward tasks. In contrast, VidMan \cite{vidman} adopts Open-Sora \citep{opensora} as its foundation for embodied video prediction, necessitating large-scale pretraining with the Open X-Embodiment dataset \cite{Open-X}. Our approach diverges by introducing optical flow as an intermediate variable to bridge natural and robot data, thereby reducing the data requirements for fine-tuning. Other methods construct world models conditioned on robot actions \cite{irasim,avid,adaworld}, yet incorporating actions as an independent modality demands extensive action-labeled trajectories.

\section{Experiments}
\label{sec:expe}

\subsection{Experiment setup}
\label{sec:expe_set}
\paragraph{Simulation Setup}
We evaluate our method on RoboTwin \cite{mu2024robotwin}, a generative digital twin framework that produces diverse expert demonstrations and realistic, interactive scenarios. RoboTwin builds a real world-aligned environment and covers common robotic manipulation skills such as picking and placing, as well as more complex dual-arm collaboration tasks. This enables a systematic assessment of the performance and generalization capability of our bimanual policy.
\paragraph{Real-World Setup and Data Collection~}
As shown in Fig.~\ref{fig:framework}(a), we built a natural furniture environment with two robotic arms mounted on either side of a desk and a front-facing camera to capture arm motions and scene context. To enable efficient data collection for dual-arm tasks, we developed a vision-based teleoperation system using Vision Pro. It offers an intuitive interface for controlling and collecting. Our VR-based system captures 6D wrist poses and finger joint positions, which are converted into robot-executable commands through a two-step coordinate transformation. The transformation aligns the Vision Pro coordinate frame with that of the robot and corrects for pose definition inconsistencies via Euler rotation. Finger distance is mapped to gripper width, allowing intuitive gripper control through hand gestures. The details can be found in Appendix~\ref{appendix:data collection}.

\paragraph{Architecture Details}
Our text-to-flow and flow-to-video models are initialized using the pretrained CogVideoX-2B model. We first fine-tune both models on a combined dataset consisting of two bimanual datasets: RDT and RoboMIND. We then further fine-tune it on downstream simulation and real-world demonstrations. All video samples are standardized to a resolution of $256 \times 256$ and a fixed length of 17 frames. For the text-to-flow model, we use FlowFormer++ \cite{flowformer++} to extract ground-truth optical flow from each video clip. More training details are provided in Appendix \ref{appendix:training_detail}.

\paragraph{Baselines}
We compare our method with three baselines: (i) Single-view based method: Diffusion Policy (DP) \cite{diffusionpolicy}. A conditional denoising diffusion model that generates actions by iteratively refining samples, enabling stable training and handling of complex, multi-modal behaviors. (ii) Multi-view based method: RDT \cite{rdt}. A large diffusion-based foundation model for bimanual manipulation with a unified action space and Transformer architecture. (iii) 3D-based method: DP3 \cite{ze20243d}. A 3D visual imitation learning method that leverages point cloud-based representations in diffusion policies. Our text-to-flow model, flow-to-video model, and RDT are trained on a mixed multi-task dataset, whereas the downstream goal-conditioned policy and other baseline models are trained on single-task datasets. For each task, we collect 100 demonstrations. We use the default D435 camera for RDT, and the L515 camera for other single-view baselines as well as for our method, since relying on single images as input is more challenging and requires richer visual information.

\subsection{Main Results in Simulation Setup}
\label{sec:sim_exp}
We evaluate each task using 10 random seeds, where the environment is randomly reset for each seed. For each seed, we perform 10 evaluation runs and report the average success rate along with the standard deviation. As shown in Table \ref{tab:robotwin}, due to the limited visual information provided by a single camera, the vanilla DP performs worse than other baselines that leverage additional visual input, such as 3D scene representations or multi-view images. However, when augmented with the ability to anticipate future states — enabled by conditioning on the goal predicted by our video prediction model — the success rate improves and even surpasses those of the stronger baselines in some tasks. This demonstrates the effectiveness of integrating video prediction models with bimanual policies.

\begin{figure*}[t]
    \centering
    \includegraphics[width=\textwidth]{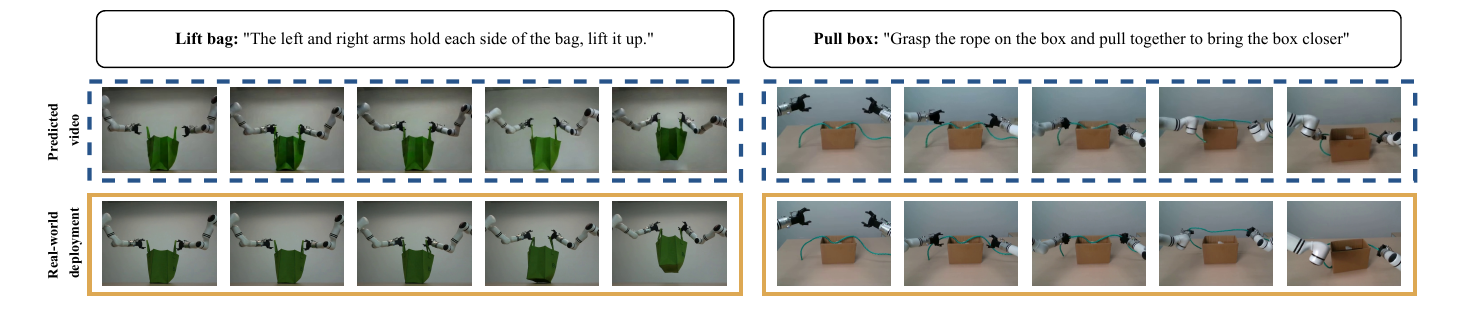}
    \vspace{-20pt}
    \caption{Visualization of the predicted video (blue dashed box) and the actual execution trajectory of the real-world dual-arm system (yellow box).}
    \label{fig:vis_realworld}
\end{figure*}

\begin{table}[t]
\vspace{5pt}
    \centering
    \caption{Results on RoboTwin. Performance comparison across different tasks and different models requiring different input type.}
    \scriptsize
    \resizebox{0.85\textwidth}{!}{\begin{tabular}{lc  lc}
        \toprule
        \textit{Put Apple Cabinet} &  & \textit{Block Handover} &  \\
        \midrule
        DP3 (Point Cloud) & 74.7 $\pm$ 42.2 & DP3 (Point Cloud) & 67.3 $\pm$ 7.0 \\
        DP3 (Point Cloud w/ color) & 97.0 $\pm$ 2.6 & DP3 (Point Cloud w/ color) & 86.0 $\pm$ 15.1 \\
        RDT (Multi-view RGB) & 22.0 $\pm$ 0.10 & 
        RDT (Multi-view RGB) & 70.0 $\pm$ 0.08 \\
        \hdashline
        DP (Single-view RGB) & 82.0 $\pm$ 0.09 & DP (Single-view RGB) & 2.0 $\pm$ 0.04 \\
        CogRobot (Single-view RGB) & 100.0 $\pm$ 0.0 & CogRobot (Single-view RGB) & 36.0 $\pm$ 0.11 \\
        \midrule
        \textit{Pick Apple Messy} &  & \textit{Container Place} &  \\
        \midrule
        DP3 (Point Cloud) & 11.7 $\pm$ 5.5 & DP3 (Point Cloud) & 89.0 $\pm$ 7.5 \\
        DP3 (Point Cloud w/ color) & 68.7 $\pm$ 6.8 & DP3 (Point Cloud w/ color) & 73.3 $\pm$ 6.5 \\
        RDT (Multi-view RGB) & 31.2 $\pm$ 0.08 & RDT (Multi-view RGB) & 34.4 $\pm$ 0.09 \\ 
        \hdashline
        DP (Single-view RGB) & 3.0 $\pm$ 0.07 & DP (Single-view RGB) & 51.0 $\pm$ 0.14 \\
        CogRobot (Single-view RGB) & 15.0 $\pm$ 0.07 & CogRobot (Single-view RGB) & 55.0 $\pm$ 0.07 \\
        \bottomrule
    \end{tabular}}
    \label{tab:robotwin}
    \vspace{-1em}
\end{table}

\subsection{Real-World Experiments}
\label{sec:real_exp}
We also evaluate our method on the real-world dual-arm system shown in Fig. \ref{fig:framework}(a). We consider two representative tasks:
(\romannumeral1) \textit{Lift bag}, a bimanual collaboration task in which both arms must simultaneously lift a bag; if either arm fails, the bag cannot be lifted.
(\romannumeral2) \textit{Pull box}, a complex multi-stage task that requires the dual arms to locate a rope, position it behind the box, and then pull the box together, presenting greater challenges for coordination and collaboration. For each task, we collect 100 human teleoperation demonstrations. To evaluate the view adaptability of our method, each task is recorded from a different camera viewpoint. We use these demonstrations to train the CogVideoX-2B model that was fine-tuned on RDT and RoboMIND, as well as the goal-conditioned diffusion policy for each task. Due to the complexity of deploying models on the real-world dual-arm system, we select the most relevant single-view model DP as our baseline. During deployment, we iteratively generate new goals by inputting the current observation into the text-to-flow model and the flow-to-video model to produce future plans. We evaluate the average success rate across 20 trials. 

\begin{wraptable}{r}{.4\textwidth}
\vspace{-1em}
    \centering
    \caption{Results on real-world tasks.}
    \vspace{1em}
    \scriptsize
    \resizebox{0.4\textwidth}{!}{\begin{tabular}{lc  lc}
        \toprule
        \textit{Lift Bag} &  & \textit{Pull Box} &  \\
        \midrule
        DP       & 0.50 & DP       & 0.05 \\
        CogRobot & 0.70 & CogRobot & 0.75 \\
        \bottomrule
    \end{tabular}}
    \label{tab:realworld}
\end{wraptable}

As shown in Table \ref{tab:realworld}, our method achieves a higher success rate compared to the vanilla DP, demonstrating the advantage of incorporating a video prediction model as a high-level planner in real-world settings. This benefit is more evident in complex tasks, where traditional low-level policies often struggle. The video model provides informative plans that effectively guide the actions of the dual-arm system to accomplish the task.

We further visualize the predicted video generated by our model and compare it with the actual execution of the dual-arm system. As shown in Fig. \ref{fig:vis_realworld}, the predicted video closely matches the real-world execution, demonstrating that our model can generating physically executable trajectories.
 

\subsection{Visualization and Ablation}
\label{sec:video_exp}
\begin{figure*}[t]
    \centering
    \includegraphics[width=\textwidth]{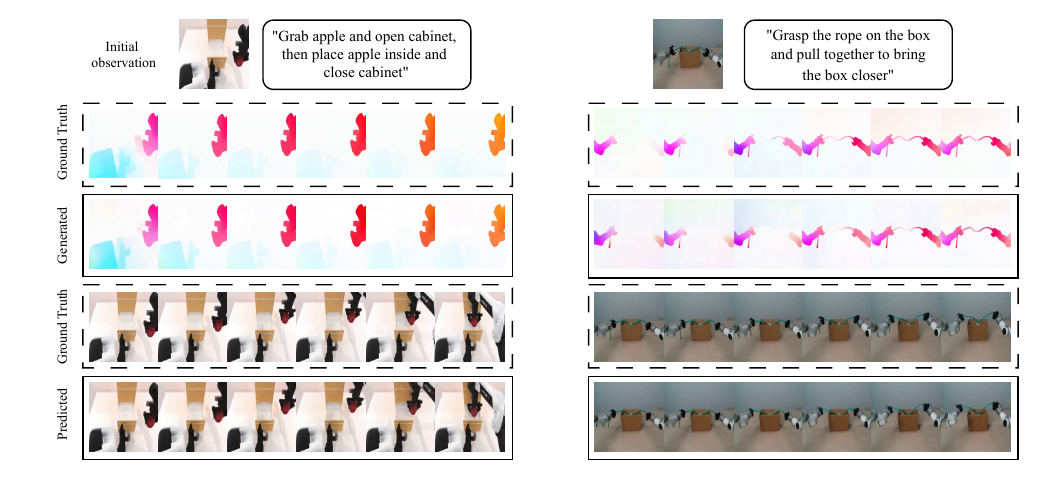}
    \vspace{-20pt}
    \caption{Visualization of the generated flow and predicted video in the RoboTwin simulation environment (left) and on the real-world Realman dual-arm system (right).}
    \label{fig:vis_flowvideo}
    \vspace{-1em}
\end{figure*}

\paragraph{Visualization} We visualize the generated optical flow and manipulation video for both simulation and real-world tasks in Fig. \ref{fig:vis_flowvideo}. Our text-to-flow model accurately reconstructs the optical flow, providing reliable motion guidance to the subsequent flow-to-video model for predicting fine-grained movements and interactions. More visualization can be found in Appendix \ref{appendix:robotwin_vis} and Appendix \ref{appendix:realman_vis}.

\paragraph{Ablation} We compare our flow-guided video prediction model with SFT variants that are directly fine-tuned on the same dataset. Specifically, we fine-tune the pre-trained CogVideoX-2B and CogVideoX-5B using the same resolution and video length on the combined datasets of RDT and RoboMIND. To evaluate video generation quality, we use the official validation set from RoboMIND and construct a separate validation set in RDT. The RoboMIND validation set contains 5,346 samples and the RDT validation set contains 1,757 samples. More details on the validation data are provided in the Appendix \ref{appendix:training_detail}. We compare the generated videos with ground-truth videos using four metrics: PSNR \cite{psnr}, SSIM \cite{ssim}, LPIPS \cite{lpips} and FVD \cite{fvd}. As shown in Table \ref{table:video_metrics}, incorporating optical flow as guidance improves the quality of the generated videos when facing with limited datasets. 

\begin{table}[h]
\centering
\caption{Video generation quality on validation datasets from  
RDT and RoboMIND using different models. Arrows indicate whether higher or lower values correspond to better performance.}
\small
\label{table:video_metrics}
\begin{tabular}{llcccc}
\toprule
\textbf{Dataset} & \textbf{Method} & \textbf{PSNR} $\uparrow$ & \textbf{SSIM} $\uparrow$ & \textbf{LPIPS} $\downarrow$ & \textbf{FVD} $\downarrow$ \\
\midrule
\multirow{3}{*}{RDT} 
  & CogVideoX-2B-SFT & 19.677 & 0.784 & 0.151 & 1222 \\
  & CogVideoX-5B-SFT & 20.502 & 0.806 & 0.132 & 1064 \\
  & CogVideoX-2B-Flow (ours) & \textbf{22.663} & \textbf{0.836} & \textbf{0.097} & \textbf{760} \\
\midrule
\multirow{3}{*}{RoboMIND} 
  & CogVideoX-2B-SFT & 18.928 & 0.821 & 0.136 & 848 \\
  & CogVideoX-5B-SFT & 19.418 & 0.837 & 0.123 & 740 \\
  & CogVideoX-2B-Flow (ours) & \textbf{21.977} & \textbf{0.864} & \textbf{0.085} & \textbf{576} \\
\bottomrule
\end{tabular}
\end{table}

\section{Conclusion}
\label{sec:conclusion}
We propose CogRobot, a novel approach for learning a generalizable bimanual policy that leverages T2V models as high-level planners, followed by a low-level policy that extracts actions from videos. We introduce a two-stage video generation training framework that incorporates optical flow as guidance to facilitate fine-grained bimanual trajectory prediction. We build a dual-arm system and use a VR-based teleoperation system for efficient data collection. Experiments in simulated environments and real-world dual-arm deployments demonstrate the effectiveness of our method. However, our method still has limitations, as each task requires a separate policy to extract actions from videos.

\clearpage

\section*{Acknowledgments}
This work is supported by the National Key Research and Development Program of China (Grant No.2024YFE0210900), the National Natural Science Foundation of China (Grant No.62306242),  the Young Elite Scientists Sponsorship Program by CAST (Grant No. 2024QNRC001), and the Yangfan Project of the Shanghai (Grant No.23YF11462200).


\small
\bibliography{main}
\bibliographystyle{unsrtnat}
\clearpage

\clearpage
\appendix
\section{Details of Data Collection}
\label{appendix:data collection}
\begin{figure*}[h]
    \centering
    \includegraphics[width=0.98\textwidth]{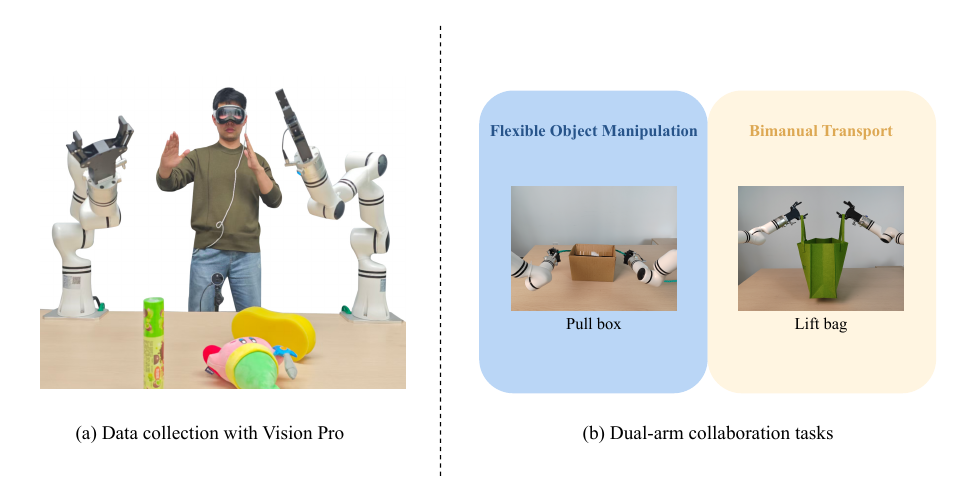}
    \caption{Visualization of the real-world experimental setup and tasks. (a) We develop a VR-based teleoperation system for efficient demonstration collection. (b) We design two types of real-world bimanual manipulation tasks, each requiring coordinated dual-arm collaboration.}
    \label{appendix:dataset}
\end{figure*}

\paragraph{Platform Settings}
To address embodiment gaps encountered when training with Internet bimanual data and to better adapt to downstream tasks, we developed a custom teleoperation system using Vision Pro for efficient collection of high-quality dual-arm manipulation data. As shown in Fig.~\ref{appendix:dataset}(a), our teleoperation system provides an intuitive interface for both controlling the robot and recording demonstrations. A RealSense 435i camera is positioned in front of the workspace to capture comprehensive single-view observations, including the movements of the robotic arms and the state of the environment. During the data collection phase, for each completed task, we record the following data: third-person videos, dual-arm joint actions, task instructions, and gripper opening degrees. Our goal is to capture detailed on-screen appearances of the robotic arms, which helps the video prediction model better anticipate the trajectories they should follow during the task.

\paragraph{Teleoperation method}
High-quality data collection is essential for bimanual policy learning, as both the fine-tuning of the video diffusion model and the training of the low-level control policy rely on a large volume of high-quality demonstrations of robots performing diverse real-world tasks. Efficiently obtaining such data presents a significant challenge, highlighting the need for a well-designed and effective teleoperation system.
However, many open-source teleoperation solutions \cite{ding2024bunny},\cite{opentelevision},\cite{he2024omnih2o} are tightly coupled with specific robotic platforms, making them time-consuming and technically challenging to adapt to other robotic arms. To address this limitation, we developed a teleoperation system tailored to our Realman dual-arm system, enabling more efficient and flexible data collection.

Using VR to teleoperate a dual-arm system provides an intuitive and efficient approach for data collection. However, fine-grained teleoperation via VR poses two main challenges: (i) transforming the wrist coordinates obtained from the VR device into a format that can be executed by the robotic arm to accurately replicate human hand movements in three-dimensional space, and (ii) applying the transformed coordinates to generate smooth and continuous trajectories that are suitable for downstream model learning.
In our coordinate transformation module, we adopt a concise implementation. The wrist pose captured by the Vision Pro during teleoperation undergoes two coordinate transformations to generate control signals that can be directly executed by the robotic arm’s end effector.

We adopt the open-source project \cite{park2024using} to extract the 6D wrist pose and finger joint positions. The wrist pose is represented by a 4\(\times\)4 homogeneous transformation matrix \( \mathbf{T}_{\text{hand}} \). To convert this pose into a format compatible with the robotic arm, we apply two additional transformation matrices, \( \mathbf{T}_1 \) and \( \mathbf{T}_2 \), as follows:
\begin{equation}
\mathbf{T}_{\text{arm}} = \mathbf{T}_1 \cdot \mathbf{T}_{\text{hand}} \cdot \mathbf{T}_2
\label{eq:transform}
\end{equation}
This transformation is performed at each timestep to generate a continuous trajectory for the robot to execute.
The left multiplication by \( \mathbf{T}_1 \)
maps the Vision Pro coordinate system into the robot's coordinate frame. Specifically, the rotation component of \( \mathbf{T}_1 \) aligns the axes of the two coordinate systems, while the translation component adjusts the origins.
The right multiplication by \( \mathbf{T}_2 \) addresses pose alignment. Since different coordinate systems define the initial pose of an object differently, the rotational components of an identical pose may vary. The multiplication by \( \mathbf{T}_2 \) applies an Euler rotation around the end-effector’s own axis to obtain the pose that matches the hand’s orientation.
After applying these coordinate transformations, the hand’s movements and rotations in all spatial directions can be accurately reproduced by the robotic arm. In addition, the distance between the index finger and thumb is normalized to a value between 0 and 1 and used to control the gripper's opening width, enabling intuitive gripper control via hand gestures.

To ensure smooth trajectories of the robotic arm in both position and velocity, we adopt a lightweight execution strategy. During data collection, the control frequency is set to 60 Hz. At each timestep, the transformed target pose is directly transmitted to the robot controller, which performs inverse kinematics to compute a candidate joint configuration. If a valid solution is found and the resulting joint angles differ only slightly from the current configuration, the command is executed immediately without additional trajectory planning.
This approach is well suited for our scenario, where frequent and real-time pose updates are required.

\section{Details of Real-world Experiments}
We consider tasks that require coordinated bimanual manipulation and cannot be accomplished by a single arm alone. Specifically, we design two categories of tasks: \textit{Flexible Object Manipulation} and \textit{Bimanual Transport}. In \textit{Flexible Object Manipulation}, the dual arms must cooperatively handle flexible objects such as ropes to achieve specific goals—for example, pulling a box closer using a rope (\textit{Pull box}). In \textit{Bimanual Transport}, the dual arms must jointly move target objects, such as lifting a bag together (\textit{Lift bag}). These tasks are designed to highlight the advantages of coordinated bimanual manipulation. Detailed task instructions are provided in Table~\ref{appendix:dataset_tab}. Visualizations for each task are presented in Fig.~\ref{appendix:dataset}(b).

\begin{table}[htbp]
\centering
\caption{Instructions for our real-world bimanual manipulation tasks.}
\label{appendix:dataset_tab}
\begin{tabularx}{.9\linewidth}{l l X}
\toprule
\textbf{Task} & \textbf{Subtask} & \textbf{Instruction} \\
\midrule
\multirow{1}{*}{Flexible Object Manipulation} 
& Pull box & \textit{Both the left and right arms simultaneously grasp the two ends of the rope on the box, place the rope behind the box, and then pull together to bring the box closer.} \\
\midrule
\multirow{1}{*}{Bimanual Transport} 
& Lift bag & \textit{The left arm grabs the left handle of the bag, and the right arm grabs the right handle. Both arms lift the bag together.} \\
\bottomrule
\end{tabularx}
\end{table}



\section{Implementation Details}
\label{appendix:training_detail}
\subsection{Architecture details}
Both our text-to-flow and flow-to-video models are based on the pre-trained CogVideoX-2B model. However, the official 2B version does not support initial observations as an additional conditioning input. Therefore, we use a community-pretrained variant provided by VideoX-Fun (\url{https://huggingface.co/alibaba-pai/CogVideoX-Fun-V1.1-2b-InP}), which supports additional input images and multiple resolutions. For ablations on directly fine-tune CogVideoX, we use the same 2B model and the 5B variant provided by VideoX-Fun (\url{https://huggingface.co/alibaba-pai/CogVideoX-Fun-V1.1-5b-InP}).

\begin{figure*}[h]
    \centering
    \includegraphics[width=0.98\textwidth]{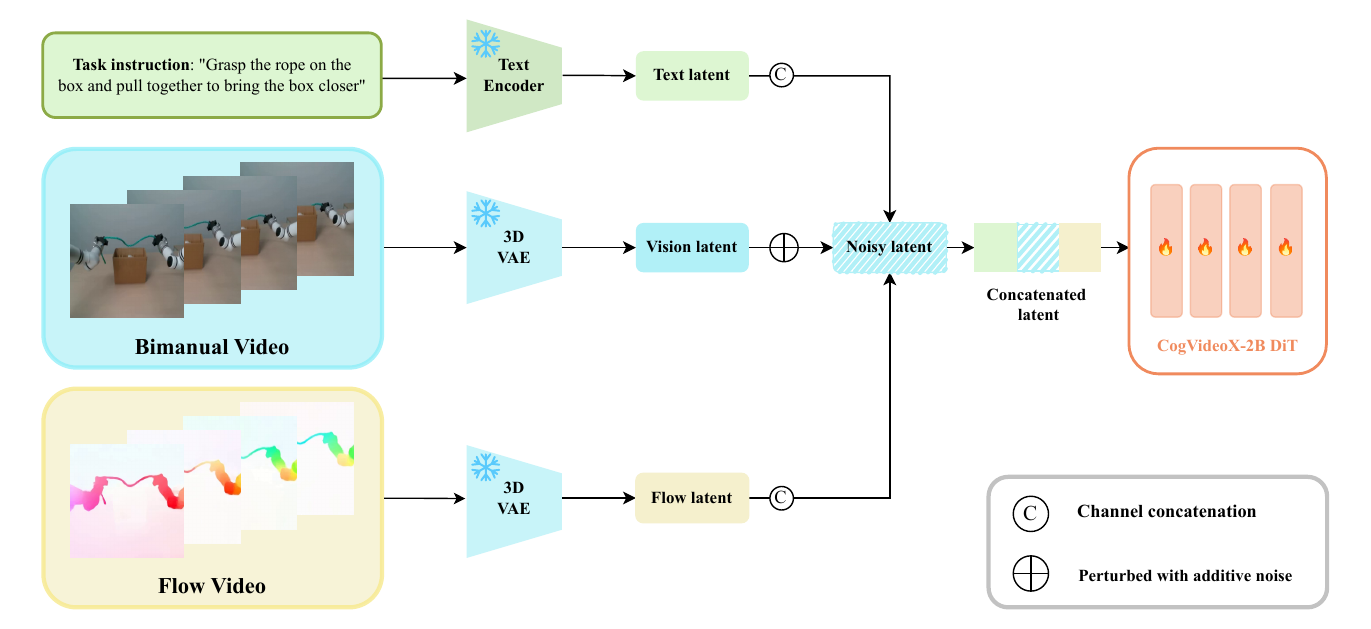}
    \caption{Architecture of the flow-to-video model. The bimanual video and the flow video are encoded using the same three-dimensional VAE, and their corresponding latents are concatenated along the channel dimension to form the input to the model.}
    \label{appendix:f2vmodel}
\end{figure*}

For the text-to-flow model, after converting the raw optical flow into flow videos, we directly fine-tune these video data in an end-to-end manner to obtain the flow generation model. During training, the VAE is kept frozen, and only the diffusion transformer is fine-tuned to predict the noise added to the original flow video.
For the flow-to-video model, the architecture is illustrated in Fig.~\ref{appendix:f2vmodel}. The flow video and the bimanual manipulation video are separately encoded using the same pre-trained 3D VAE from CogVideoX, which remains frozen during fine-tuning. Noise is then added to the video latent following the standard video prediction process. The resulting noisy video latent is concatenated along the channel dimension with the text embedding, which is processed by a pre-trained text encoder, and the flow latent. The final input to the flow-to-video model is thus a joint latent that integrates language, video, and flow information. The model is trained to predict the noise added to the video latent.

For the low-level goal-conditioned diffusion policy, we adpot the implementation from RoboTwin (\url{https://github.com/TianxingChen/RoboTwin}). We introduce an additional ResNet-18 model to encode the goal image, and its feature representation is concatenated with the feature extracted by another ResNet-18 model that processes the history of observations.

\subsection{Datasets}
We fine-tune both our text to flow model and flow to video model on a combination of existing open source bimanual datasets and self-collected datasets from simulation and real world environments. For the internet bimanual datasets, we use RDT~\cite{rdt} and RoboMIND~\cite{robomind}. RDT is a bimanual manipulation dataset consisting of more than 300 diverse tasks, collected using the ALOHA system~\cite{mobilealoha}. RoboMIND is a cross embodiment dataset, from which we use only the bimanual AgileX subset that also includes a variety of tasks. To evaluate the quality of video generation, we construct a separate test set by excluding certain samples from the training data. Specifically, we exclude $5\%$ of the episodes from RDT as a validation set. For RoboMIND, we use the official validation set as the test set.
For the downstream simulation and real world experiments, we collect 100 demonstrations for each task. Specifically, for RoboTwin, we use the official data collection script to collect 100 successful trajectories per task. For Realman, we use the VR-based teleoperation system to collect 100 human demonstrations per task.

\paragraph{Data pre-processing} We resize all videos in the datasets to a resolution of $256 \times 256$ and divide them into video clips with a fixed length of 17 frames. Since the movements in bimanual trajectories are often slight due to the need for precise manipulation, sampling all frames from the original videos would result in many consecutive frames with little visual change. To address this, we down-sample the videos by skipping frames during the clipping process. For each dataset, we use a fixed sampling stride to determine the start frame index, and apply a larger stride when clipping the validation set to reduce evaluation cost. For the text to flow model, we use FlowFormer++~\cite{flowformer++} to generate the ground truth optical flow. Specifically, we estimate the optical flow between frame 0 and each subsequent frame $i$, where $i \in [1, 2, \dots, 16]$, and concatenate the resulting flow maps along the temporal dimension to form a flow sequence. After converting the flow sequence into a flow video, we prepend a pure white image at the beginning to match the total video length to 17 frames. We also filter out frames with low optical flow magnitude to further reduce the impact of still frames. The resulting number of samples for each dataset, along with the sampling stride and frame skipping interval, is presented in Table \ref{appendix:data_statistics}.

\begin{table}[h]
\centering
\caption{Dataset statistics and pre-processing parameters. The validation set is used to evaluate the quality of video generation. We sample video clips using a fixed stride and apply frame skipping to reduce the presence of still frames caused by small motion in bimanual trajectories.}
\small
\label{appendix:data_statistics}
\begin{tabular}{llccc}
\toprule
\textbf{Dataset} & \textbf{Data Split} & \textbf{Sample stride} & \textbf{Down-sample interval} & \textbf{Samples} \\
\midrule
\multirow{2}{*}{RDT} 
  & Training & 4 & 2 & 157,912  \\
  & Validation & 16 & 2 & 1,757  \\
\midrule
\multirow{2}{*}{RoboMIND} 
  & Training & 4 & 2 & 198,789  \\
  & Validation & 16 & 2 & 5,346  \\
\midrule
\multirow{1}{*}{RoboTwin} 
  & Training & 1 & 4 & 106,771  \\
\midrule
\multirow{1}{*}{Realman} 
  & Training & 1 & 4 & 12,866  \\
\bottomrule
\end{tabular}
\end{table}

\subsection{Training details}
For both the text-to-flow and flow-to-video models, fine tuning is performed in two stages. We first fine tune the CogVideoX model on the combined dataset of RDT and RoboMIND. The text-to-flow and flow-to-video models are fine tuned separately: the text-to-flow model is trained using the ground truth flow video as the target, while the flow-to-video model uses the ground truth flow video as input. In the second stage, the fine-tuned text-to-flow and flow-to-video models are used as base models and further fine-tuned on specific downstream datasets: RoboTwin or Realman.

For all models, we use AdamW~\cite{adamw} as the optimizer with a learning rate of $2 \times 10^{-5}$ and apply learning rate warm-up during the first 100 steps. For fine-tuning on the RDT and RoboMIND datasets, the text-to-flow model is fine-tuned for 20k steps, and the flow-to-video model for 30k steps. For the RoboTwin dataset, both models are fine-tuned for 15k steps. For the Realman dataset, both models are fine-tuned for 5k steps. All models are fine-tuned using four H100 80GB GPUs, with a batch size of 128 per GPU.

\subsection{Details of baselines}
We describe the details of baselines used for comparison in our simulation and real-world experiments. For simulation experiment, we consider following baselines and use the implementation from \url{https://github.com/TianxingChen/RoboTwin}:
\begin{itemize}[leftmargin=*]
\item \textbf{Diffusion Policy (DP).} 
A visual imitation learning algorithm that employs a diffusion model to generate future action sequences conditioned on past observation histories. The original DP is designed for single-arm manipulation. We extend it to support dual-arm actions by modifying the action space to accommodate both arms. We use the default hyperparameters provided in the official RoboTwin implementation.
\item \textbf{3D Diffusion Policy (DP3).}
An improved version of DP that incorporates 3D visual representations by using a point cloud encoder to transform visual observations into feature embeddings. Similar to DP, we adapt the action space to support dual-arm systems by modifying the action dimension accordingly. We use the default hyperparameters provided in the official RoboTwin implementation.
\item \textbf{RDT.}
A 1B-parameter diffusion-based bimanual policy pre-trained on a large-scale Internet dataset of robotic arm manipulation. It introduces a unified action space to standardize the action dimensions and types across different robotic arms. The policy generates the unified action sequence conditioned on three-view image observations: one from a third-person camera and two from wrist-mounted cameras on each arm. We use the default hyperparameters provided in the official RoboTwin implementation.
\end{itemize}

For real-world experiments, we primarily consider DP, as it aligns more closely with our setup (i.e., single-view RGB input). We apply the same hyperparameters used in the simulation experiments when training on our collected dataset.

\clearpage
\section{More Visualization on Robotwin}
\label{appendix:robotwin_vis}

\begin{figure*}[h]
    \centering
    \includegraphics[width=0.98\textwidth]{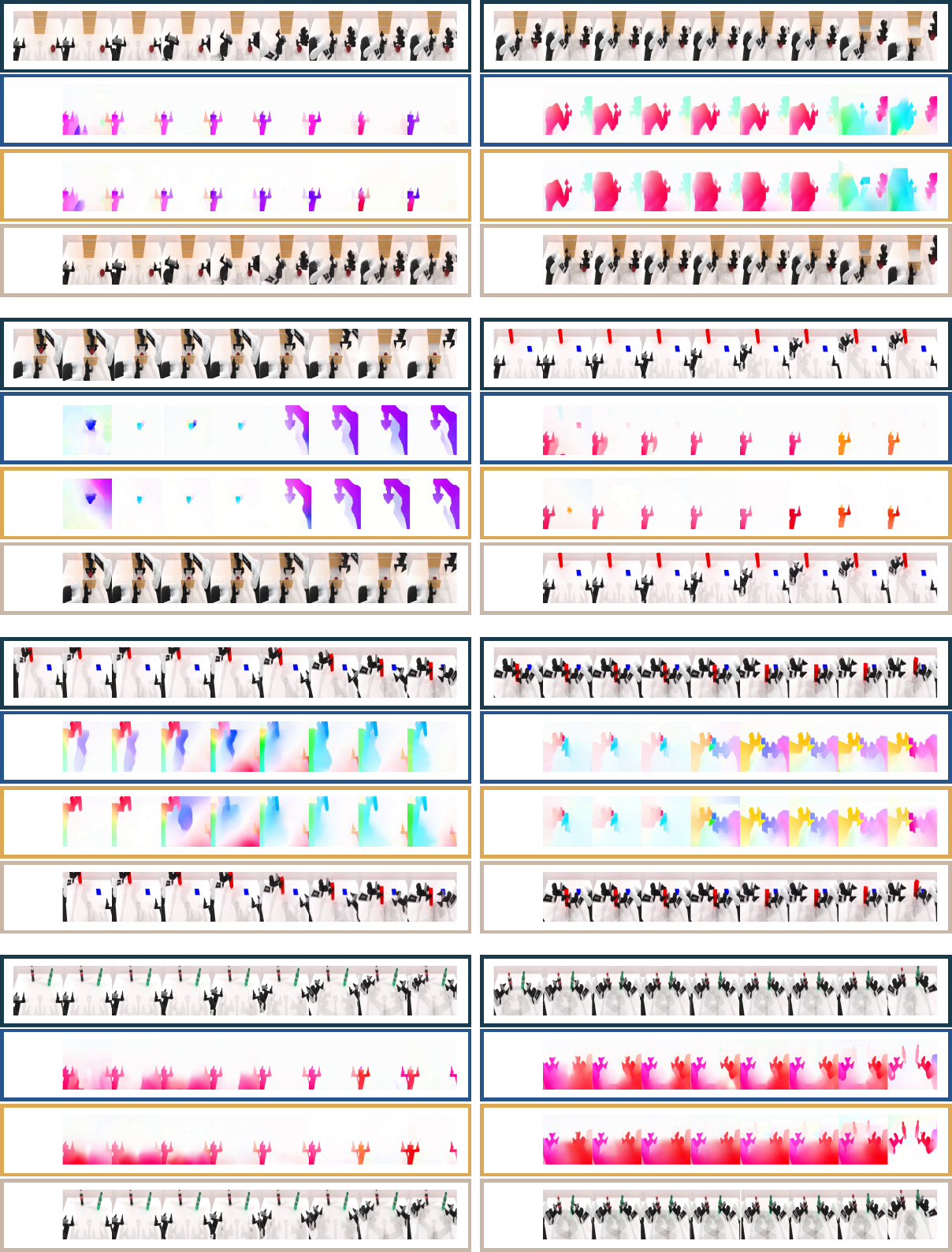}
    \caption{More visualization of the generated flow and predicted video in the RoboTwin simulation environment. The \textcolor[HTML]{183B4E}{black box} indicates the ground truth video. The \textcolor[HTML]{27548A}{blue box} shows the ground truth optical flow represented in RGB. The \textcolor[HTML]{DDA853}{yellow box} shows the flow generated by the text-to-flow model. The \textcolor[HTML]{C8B6A6}{brown box} shows the video predicted by the flow-to-video model using the generated flow.}
\end{figure*}

\clearpage
\section{More Visualization on Realman}
\label{appendix:realman_vis}

\begin{figure*}[h]
    \centering
    \includegraphics[width=0.98\textwidth]{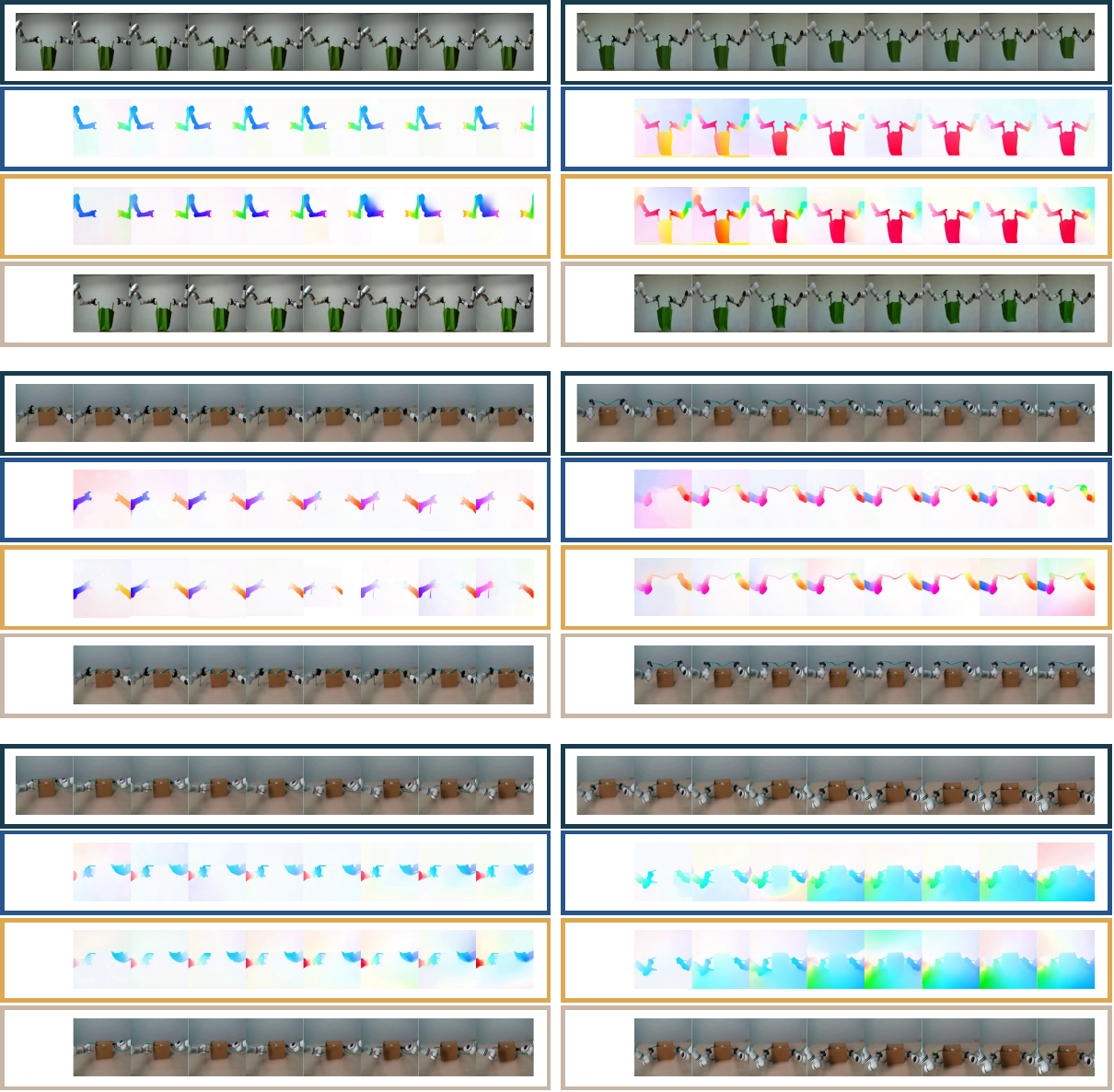}
    \caption{More visualization of the generated flow and predicted video for the real-world Realman dual-arm system. The \textcolor[HTML]{183B4E}{black box} indicates the ground truth video. The \textcolor[HTML]{27548A}{blue box} shows the ground truth optical flow represented in RGB. The \textcolor[HTML]{DDA853}{yellow box} shows the flow generated by the text-to-flow model. The \textcolor[HTML]{C8B6A6}{brown box} shows the video predicted by the flow-to-video model using the generated flow.}
\end{figure*}


\end{document}